\documentclass[sigconf]{acmart}
\AtBeginDocument{%
  }

\copyrightyear{2026}
\acmYear{2026}
\setcopyright{cc}
\setcctype{by}
\acmConference[SIGIR '26]{Proceedings of the 49th International ACM SIGIR Conference on Research and Development in Information Retrieval}{July 20--24, 2026}{Melbourne, VIC, Australia}
\acmBooktitle{Proceedings of the 49th International ACM SIGIR Conference on Research and Development in Information Retrieval (SIGIR '26), July 20--24, 2026, Melbourne, VIC, Australia}
\acmDOI{10.1145/3805712.3809540}
\acmISBN{979-8-4007-2599-9/2026/07}

\usepackage{graphicx} 
\usepackage{booktabs}
\usepackage[dvipsnames,svgnames]{xcolor}
\usepackage{multirow}
\usepackage[compatibility=false]{caption}
\usepackage{makecell}
\usepackage{bm}
\usepackage{algorithm}  
\usepackage{algorithmicx}  
\usepackage{algpseudocode} 
\usepackage{amsmath}

\usepackage{amsfonts}
\usepackage{tcolorbox}
\tcbuselibrary{breakable}
\usepackage{balance}

\definecolor{cvprblue}{rgb}{0.21,0.49,0.74}
\definecolor{cYellow}{HTML}{FFFFCC}
\definecolor{cRed}{HTML}{FFCCCC} 
\definecolor{cGrey}{HTML}{F3F7F2} 
\definecolor{cGreen}{HTML}{339933}

\usepackage[dvipsnames]{xcolor}
\usepackage{colortbl}
\usepackage{subfig}
\usepackage{pifont}
\usepackage{listings}
\lstdefinelanguage{json}{
    basicstyle=\small\ttfamily,
    columns=fullflexible,
    showstringspaces=false,
    commentstyle=\color{gray}\upshape,
    morestring=[b]",
    morestring=[d]',
    morestring=[s]{`}{`},
    morecomment=[l]{//},
    morecomment=[s]{/*}{*/},
    morekeywords={true,false,null},
    keywordstyle=\color{blue}\bfseries,
    stringstyle=\color{black},
    breaklines=true,
    breakatwhitespace=true,
    literate=
     *{:}{{{\color{gray}{:}}}}{1}
      {,}{{{\color{gray}{,}}}}{1}
      {\{}{{{\color{gray}{\{}}}}{1}
      {\}}{{{\color{gray}{\}}}} }{1}
      {[}{{{\color{gray}{[}}}}{1}
      {]}{{{\color{gray}{]}}}}{1},
}
\definecolor{DarkGreen}{rgb}{0.0, 0.4, 0.0} 

\begin{document}

\title{Chain of Evidence: Pixel-Level Visual Attribution for Iterative Retrieval-Augmented Generation}

\author{Peiyang Liu}
\affiliation{%
  \institution{National Engineering Research Center for Software Engineering, Peking University}
  \city{Beijing}
  \country{China}}
\email{liupeiyang@pku.edu.cn}

\author{Ziqiang Cui}
\affiliation{%
  \institution{City University of Hong Kong}
  \city{Hong Kong SAR}
  \country{China}}
\email{ziqiang.cui@my.cityu.edu.hk}

\author{Xi Wang}
\affiliation{%
  \institution{Peking University}
  \city{Beijing}
  \country{China}}
\email{wangxi5629@pku.edu.cn}

\author{Di Liang}
\affiliation{%
  \institution{Tencent Technology}
  \city{Beijing}
  \country{China}}
\email{liangd17@fudan.edu.cn}

\author{Wei Ye}
\authornote{Corresponding Author}
\affiliation{%
  \institution{National Engineering Research Center for Software Engineering, Peking University}
  \city{Beijing}
  \country{China}}
\email{wye@pku.edu.cn}

\renewcommand{\shortauthors}{Peiyang Liu, Ziqiang Cui, Xi Wang, Di Liang, and Wei Ye}

\begin{abstract}
Iterative Retrieval-Augmented Generation (iRAG) has emerged as a powerful paradigm for answering complex multi-hop questions by progressively retrieving and reasoning over external documents. However, current systems predominantly operate on parsed text, which creates two critical bottlenecks: (1) \textit{Coarse-grained attribution}, where users are burdened with manually locating evidence within lengthy documents based on vague text-level citations; and (2) \textit{Visual semantic loss}, where the conversion of visually rich documents (e.g., slides, PDFs with charts) into text discards spatial logic and layout cues essential for reasoning. To bridge this gap, we present \textbf{Chain of Evidence (CoE)}, a retriever-agnostic visual attribution framework that leverages Vision-Language Models to reason directly over screenshots of retrieved document candidates. CoE eliminates format-specific parsing and outputs precise bounding boxes, visualizing the complete reasoning chain within the retrieved candidate set. We evaluate CoE on two distinct benchmarks: \textbf{Wiki-CoE}, a large-scale dataset of structured web pages derived from 2WikiMultiHopQA, and \textbf{SlideVQA}, a challenging dataset of presentation slides featuring complex diagrams and free-form layouts. Experiments demonstrate that fine-tuned Qwen3-VL-8B-Instruct achieves robust performance, significantly outperforming text-based baselines in scenarios requiring visual layout understanding, while establishing a retriever-agnostic solution for pixel-level interpretable iRAG.
Our code is available at \url{https://github.com/PeiYangLiu/CoE.git}.
\end{abstract}

\ccsdesc[500]{Information systems~Question answering}
\ccsdesc[500]{Information systems~Multimedia and multimodal retrieval}

\keywords{Multihop Question Answering, Retrieval Augmented Generation, Source Attribution}

\maketitle

\section{Introduction}
\label{sec:intro}

Large Language Models (LLMs) \cite{achiam2023gpt,bai2023qwen,liu2024deepseek,li2026instructiondataselectionanswer,li2026dataselectionmultiturndialogue} have revolutionized information seeking and broad retrieval applications \cite{mu2026masked, xing2025reg4rec, li2024category, li2026cpgrec+}, yet they remain prone to hallucinations and struggle with outdated parametric knowledge \citep{rawte2023survey,ji2023survey,huang2025survey}. Retrieval-Augmented Generation (RAG) mitigates these issues by grounding responses in external corpora, thereby enhancing factual accuracy \citep{lewis2020retrieval,jiang2023active,yu2024rankrag,xiong2024benchmarking,amugongo2025retrieval,li2026retrievalgenerationunifiedframework,DBLP:conf/aaai/LiTXCZY26}. To handle complex queries requiring synthesized knowledge, iRAG systems have been developed to perform multi-step retrieval and reasoning \citep{DBLP:conf/acl/TrivediBKS23,DBLP:conf/iclr/AsaiWWSH24,DBLP:conf/acl/SuTA0024,DBLP:conf/acl/0002QPCHL0L25,DBLP:journals/corr/abs-2501-14342}. For example, answering ``Which university did the director of the film \textit{Inception} attend?'' requires identifying the director (Christopher Nolan) and then retrieving his biography, a dependency chain that single-step RAG often fails to resolve \cite{DBLP:conf/acl/FangMM25}.

Despite iRAG's success on textual benchmarks \citep{DBLP:conf/coling/HoNSA20}, a critical disconnect remains between \textit{generation} and \textit{verification} in high-stakes domains like healthcare, finance, and law \cite{ng2025rag,wang2025financial,wiratunga2024cbr}, where verifying \textit{why} an answer was generated is essential \citep{chander2025toward}. While recent citation-based approaches \citep{DBLP:conf/emnlp/GaoYYC23,DBLP:conf/naacl/Ye0AP24,DBLP:conf/acl/MaZKZCL25} attempt to bridge this gap, they prove inadequate for diverse, visually rich real-world documents. We identify three key challenges in iRAG attribution:

    1. \textbf{The Verification Bottleneck:} Existing systems typically provide coarse-grained, text-level citations (e.g., ``[Source: Doc-1]''). In multi-hop scenarios involving multiple documents, this forces users to manually scan hundreds of pages to locate the specific sentence supporting a claim. This high cognitive load undermines the utility of the attribution itself.
    
    2. \textbf{Information Loss in Text Conversion:} Real-world knowledge is rarely just plain text. It resides in PDFs, presentation slides, and web reports containing charts, diagrams, and complex layouts. Traditional RAG pipelines rely on OCR or text parsing \cite{castro2003html} to linearize these documents. This process inevitably destroys semantic information encoded in visual structures, such as the trend in a bar chart, the causal flow in a diagram, or the hierarchy in a slide layout. For such documents, a text-based citation is not just hard to verify; it is often fundamentally insufficient because the evidence exists in the \textit{visual} relationship between elements, not in the text.
    
    3. \textbf{Opaque Reasoning Chains:} Unlike single-step retrieval, iRAG involves a trajectory of decisions. Users need to understand not just the final evidence, but the \textit{chain of evidence}: how one intermediate piece of evidence (e.g., identifying an entity) guides the selection of the next document from the candidate set. Current methods lack a unified mechanism to visualize this cross-document reasoning path.

\begin{figure*}
	    \centering
	    \includegraphics[scale=0.35]{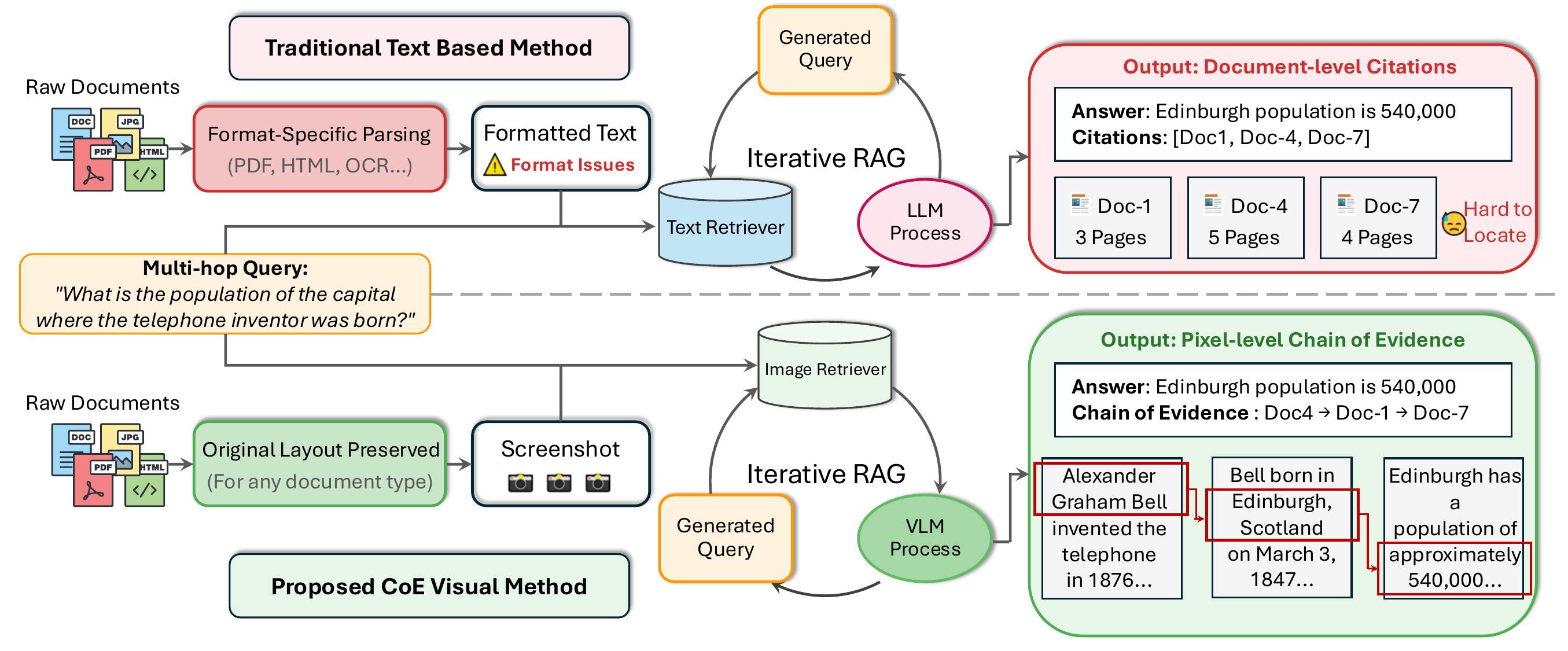}
        \vspace{-4mm}
        \caption{
        Comparison between traditional text based method and our proposed CoE visual method. CoE directly pinpoints the chain of evidences of the answer for user query in the original document with bounding boxes.
        }
	    \label{fig:overview}
\end{figure*}

To address these limitations, we propose \textbf{Chain of Evidence (CoE)}, a novel visual attribution framework that fundamentally reimagines iRAG by operating directly on document screenshots. Driven by the advancements in Vision-Language Models (VLMs) and multimodal retrieval \cite{zhu2023minigpt,zhang2024vision,guo2024regiongpt,bordes2024introduction,shinde2025survey,wei2025deepseek, INTENT, REFINE, HINT, li2026aimcotactiveinformationdrivenmultimodal}, CoE bypasses brittle text parsing pipelines. Instead, it takes visual document candidates from a retriever and generates precise bounding boxes $[(x_1, y_1), (x_2, y_2)]$ that pinpoint evidence regions, whether they are text paragraphs, table cells, or visual diagrams.
As illustrated in Figure~\ref{fig:overview}, CoE transforms the ``black box'' of multi-hop reasoning into a transparent, verifiable visual process. By grounding answers in pixel coordinates, we provide users with an immediate visual verification mechanism, significantly reducing the effort required to validate complex reasoning chains.

To rigorously evaluate CoE across different levels of visual complexity, we introduce a dual-benchmark evaluation strategy. First, we construct \textbf{Wiki-CoE}, a large-scale dataset derived from 2WikiMultiHopQA featuring 70,418 questions with bounding box annotations on structured Wikipedia layouts. Second, to challenge the model with complex, free-form visual reasoning, we incorporate \textbf{SlideVQA} \cite{SlideVQA2023}, a dataset of presentation slides where evidence is often embedded in charts, arrows, and non-linear layouts.

Our contributions are as follows:

    1. We formalize the \textbf{Chain of Evidence} problem for iRAG, proposing a visual-first framework that provides pixel-level source attribution and eliminates the need for format-specific document parsing.
    
    2. We demonstrate that visual grounding is not merely an interpretability feature but a reasoning necessity for complex documents. On the \textbf{SlideVQA} dataset, where text-based baselines fail due to layout information loss, CoE maintains robust performance by preserving visual semantics.
    
    3. We release \textbf{Wiki-CoE}, the first large-scale benchmark for multi-hop visual evidence localization, alongside our fine-tuned Qwen3-VL-8B-Instruct model.
    
    4. Extensive experiments show that CoE achieves 80.4\% evidence localization accuracy on Wiki-CoE and significantly outperforms text-based baselines on SlideVQA, offering a practical solution for trustworthy and interpretable AI systems.

\section{Related Work}

\subsection{Iterative Retrieval-Augmented Generation}
While foundational RAG systems and dense retrieval techniques demonstrated the efficacy of augmenting generation with retrieved passages \citep{zhao2024retrieval,gao2023retrieval, liu2025queries, liu2021improving, liu2021quadrupletbert, liu2021distilling}, they often struggle with complex queries requiring multi-step reasoning. Iterative RAG (iRAG) addresses this by performing multi-turn retrieval. Recent advancements focus on optimizing the retrieval process: \citet{jeong2024adaptive} proposed adaptive strategies to dynamically control retrieval frequency, while \citet{zhang2024raft} introduced retrieval-aware fine-tuning to enhance context utilization. To improve reasoning trajectories, \citet{pan2024chain} utilized explicit action chains, recent works explored synthesizing reasoning paths \cite{liu2026learningcontrastssynthesizingreasoning}, and \citet{DBLP:conf/acl/FangMM25} employed knowledge triples for active retrieval, achieving state-of-the-art performance on textual benchmarks like 2WikiMultiHopQA \citep{DBLP:conf/coling/HoNSA20}. Despite these successes, existing iRAG systems predominantly rely on parsed text, discarding visual layout cues and providing only coarse-grained citations.

\subsection{Source Attribution in LLMs}
Verifiability, along with data integrity and security, is critical for trustworthy AI \cite{liu2023retrieval, liu2024unsupervised, liu2025stole, liu2020not, liu2022label}. \citet{rashkin2023measuring} established the Attributable to Identified Sources (AIS) framework to evaluate whether generated content is supported by external evidence. Subsequent works have integrated attribution objectives into model training, either for specific QA tasks \citep{bohnet2022attributed} or during the pretraining phase \citep{khalifa2024source}. However, these approaches typically output text-level citations, forcing users to manually locate evidence within documents. Recently, VISA \citep{DBLP:conf/acl/MaZKZCL25} shifted the paradigm towards visual attribution, pinpointing evidence in single-step retrieval scenarios. Our work extends this visual grounding to multi-hop visual reasoning under a retriever-agnostic top-5 candidate setting, establishing a complete chain of visual evidence across multiple documents.

\section{Wiki-CoE Dataset}
\label{sec:wiki_coe}
\subsection{Motivation and Design Principles}

\begin{figure*}
	    \centering
	    \includegraphics[scale=0.38]{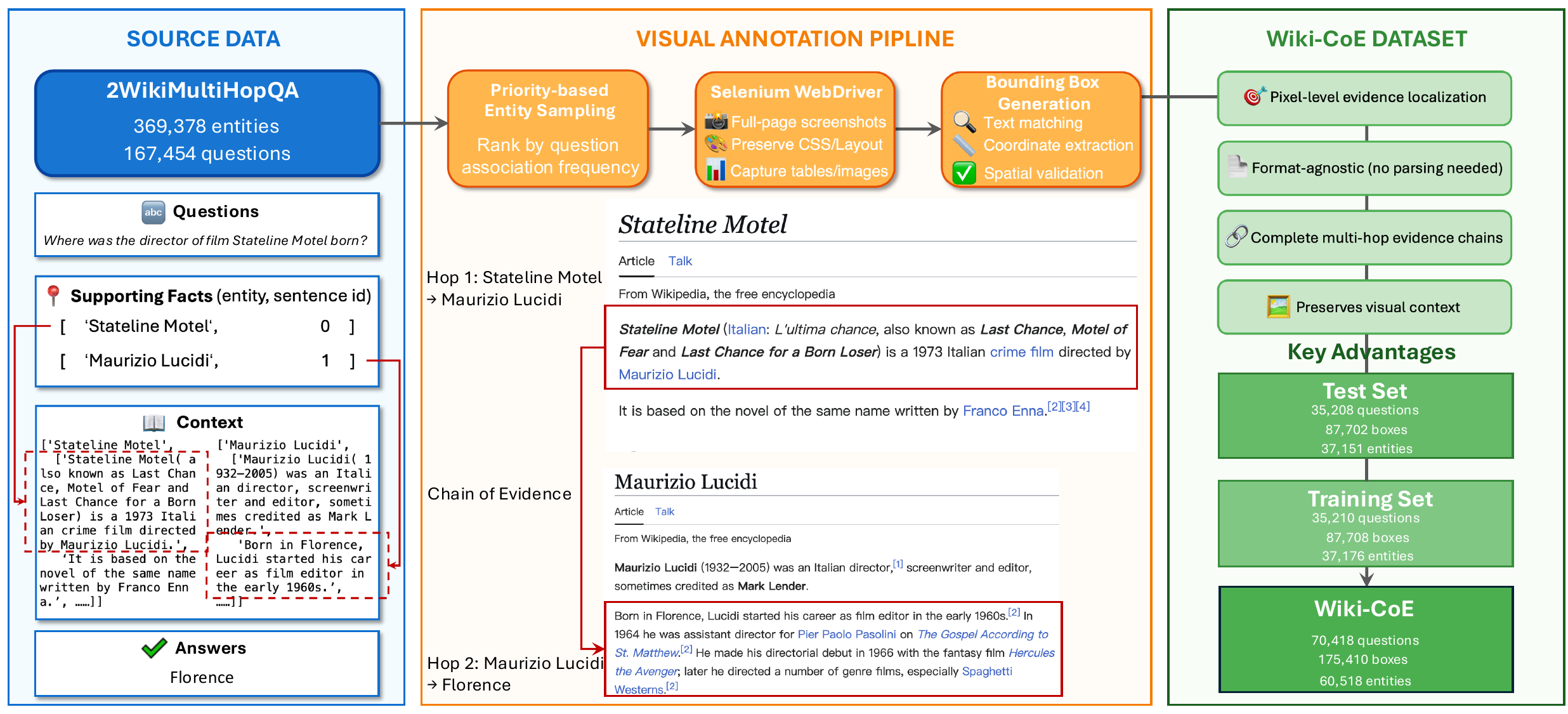}
        \vspace{-3mm}
        \caption{
        The pipline of generating our Wiki-CoE dataset.
        }
	    \label{fig:data_pipline}
\end{figure*}

Existing multi-hop QA datasets provide textual annotations but lack visual grounding essential for evaluating pixel-level attribution. While 2WikiMultiHopQA \citep{DBLP:conf/coling/HoNSA20} offers supporting facts as sentence-level annotations, these cannot directly translate to visual evidence in rendered documents where layout, formatting, and visual elements play crucial roles. As shown in Figure \ref{fig:data_pipline}, Wiki-CoE bridges this gap by providing the first large-scale benchmark with bounding boxes for visual evidence localization in multi-hop reasoning.

Our dataset design follows three principles: (1) \textbf{Visual Fidelity}: preserve original Wikipedia \cite{glott2010wikipedia} layouts including tables, infoboxes, and images that are often critical for answering questions; (2) \textbf{Evidence Completeness}: retain examples whose evidence chains can be mapped to visual bounding boxes; (3) \textbf{Scalability}: prioritize high-impact entities to maximize dataset coverage while managing computational resources.

\subsection{Dataset Construction}

Wiki-CoE extends 2WikiMultiHopQA through a systematic visual annotation pipeline:

\paragraph{Visual Document Collection.} 
We employ Selenium WebDriver \cite{garcia2022hands} to capture high-resolution screenshots of Wikipedia pages, preserving their native rendering with full CSS styling \cite{duckett2011html}, images, and interactive elements. Given the computational intensity of crawling all Wikipedia entities from the original dataset, we implement a priority-based sampling strategy. Entities are ranked by their question association frequency, the number of distinct questions requiring that entity as evidence. This ensures maximum question coverage with limited resources.

\paragraph{Bounding Box Annotation.} 
We leverage the supporting facts annotations from 2WikiMultiHopQA, which identify specific sentences serving as evidence. For each supporting fact $(entity, sentence\_id)$ pair, we:
\begin{enumerate}
    \item Extract rendered text-bearing elements and line rectangles from the live Wikipedia page, including paragraphs, list items, table cells, captions, and infobox-adjacent text.
    \item Match each supporting sentence to a rendered element using exact matching when possible and token/character-overlap similarity otherwise, then generate a bounding box $[(x_1, y_1), (x_2, y_2)]$ in screenshot pixel coordinates.
    \item Clip and validate boxes against the screenshot frame so that invalid, empty, or out-of-bounds evidence regions are discarded.
\end{enumerate}

\paragraph{Quality Assurance.}
Our construction pipeline incorporates multiple quality filters:

    1. \textbf{High Quality Texts}: The questions and answers in 2WikiMultiHopQA are human-judged, we consider this dataset a high-quality, supervised dataset with Wikipedia webpage.
    
    2. \textbf{Crawling Validation}: Screenshots are kept only when the rendered page loads successfully and the captured image has a valid size.
    
    3. \textbf{Annotation Verification}: Bounding boxes undergo automatic validation ensuring positive area, in-frame coordinates, and sufficient textual correspondence with the original supporting facts.
    
    4. \textbf{Noise Filtering}: We remove or repair instances where evidence cannot be matched to valid rendered regions with sufficient confidence, so each released example contains in-frame evidence boxes.

The released screenshot pool contains 76,000 rendered Wikipedia pages. After strict quality filtering, Wiki-CoE contains 70,418 multi-hop questions, partitioned into train (35,210) and test (35,208) splits at the entity-chain level so that no entity chain appears in both sides. The cleaned benchmark references 60,518 unique evidence screenshots across the two splits.
The questions include the following types:  
1. Comparison: Comparing the differences between two entities regarding a specific attribute.  
2. Inference: Reasoning based on logical rules from the knowledge base.  
3. Compositional: Requiring the integration of multiple independent facts to answer.  
4. Bridge comparison: A complex form of comparative questions that requires first identifying a ``bridging'' entity before the comparison can be made.
Detailed dataset statistics can be found in Table \ref{tab:dataset_stats}.

\begin{table}[t]
\centering
\small
\begin{tabular}{lrrr}
\toprule
\textbf{Statistic} & \textbf{Train} & \textbf{Test} & \textbf{Total} \\
\midrule
\multicolumn{4}{l}{\textit{Question and Answer}} \\
\quad Questions & 35,210 & 35,208 & 70,418 \\
\quad Avg. question length & 13.06 & 13.03 & 13.05 \\
\quad Avg. answer length & 1.99 & 1.99 & 1.99 \\
\midrule
\multicolumn{4}{l}{\textit{Evidence Chain}} \\
\quad Unique evidence screenshots & 37,176 & 37,151 & 60,518 \\
\quad Total bounding boxes & 87,708 & 87,702 & 175,410 \\
\quad Avg. boxes & 2.49 & 2.49 & 2.49 \\
\quad 2 hops (\%) & 75.5 & 75.5 & 75.5 \\
\quad 4 hops (\%) & 24.5 & 24.5 & 24.5 \\
\midrule
\multicolumn{4}{l}{\textit{Question Type Distribution}} \\
\quad Bridge comparison (\%) & 24.5 & 24.5 & 24.5 \\
\quad Comparison (\%) & 16.3 & 16.2 & 16.3 \\
\quad Inference (\%) & 2.3 & 2.2 & 2.3 \\
\quad Compositional (\%) & 56.9 & 57.0 & 56.9 \\
\bottomrule
\end{tabular}
\caption{
Comprehensive statistics of the cleaned Wiki-CoE release. Train/test are split at the entity-chain level, so no entity chain appears in both sides. For unique evidence screenshots, the total column is deduplicated across train and test splits.
}
\label{tab:dataset_stats}
\end{table}

\section{Methodology}

\subsection{Problem Formulation}

We formalize the Chain of Evidence (CoE) task as a structured multi-modal reasoning problem over visual documents. Let $\mathcal{Q}$ denote the query space and $\mathcal{C} = \{d_1, d_2, ..., d_N\}$ represent a corpus of documents. In traditional text-based iRAG, each document $d_i$ exists as parsed text $d_i^{\text{text}}$. Our visual paradigm fundamentally reimagines this representation: each document is captured as a screenshot image $d_i^{\text{vis}} \in \mathbb{R}^{H \times W \times 3}$, preserving its native visual presentation including layout, formatting, and graphical elements.

Given a multi-hop query $q \in \mathcal{Q}$, an upstream retriever provides a candidate set $\mathcal{D}^{\text{cand}}=\{d_1,\ldots,d_k\}\subseteq\mathcal{C}$. Our objective is to learn a function $\Phi: \mathcal{Q} \times \mathcal{D}^{\text{cand}} \rightarrow \mathcal{A} \times \mathcal{E}$ that maps the query and candidate screenshots to both an answer $a \in \mathcal{A}$ and a chain of evidence $e \in \mathcal{E}$, where:

\begin{equation}
\mathcal{E} = \bigcup_{t=1}^{T} \left\{ (d_t^*, \mathcal{B}_t) : d_t^* \in \mathcal{D}^{\text{cand}}, \mathcal{B}_t \subseteq \mathbb{R}^4 \right\}.
\end{equation}
Here, $T$ denotes the number of reasoning hops, $d_t^*$ represents the pivotal document selected at hop $t$, and $\mathcal{B}_t = \{b_{t,1}, ..., b_{t,K_t}\}$ contains $K_t$ bounding boxes, where each $b_{t,k} = [x_1^{(k)}, y_1^{(k)}, x_2^{(k)}, y_2^{(k)}]$ delineates a rectangular region containing evidence within $d_t^*$.

\subsection{Retriever-Agnostic Candidate Reasoning}

CoE is not designed as a replacement for a specific retriever. Instead, it assumes a generic upstream retriever that returns a top-$k$ candidate set, and focuses on selecting, ordering, and grounding the evidence contained in those candidates. This makes the method compatible with lexical, dense, hybrid, or visual retrievers without introducing retriever-specific parameters into the CoE model.

In our experiments, we simulate this interface by constructing candidate sets from the gold evidence documents plus distractors. For SlideVQA, distractors are sampled from the same slide deck so that non-evidence candidates are visually and topically plausible. For Wiki-CoE, distractors are sampled from the available Wikipedia screenshot pool. Candidate order is shuffled in the top-5 setting, so the model cannot rely on fixed positions and must output the selected candidate image identifiers explicitly.

\subsection{Chain-Structured Evidence Generation}

Given the query and all candidate screenshots, CoE generates the complete evidence chain in a single autoregressive pass. Each candidate screenshot is labeled as \texttt{img\_0}, \texttt{img\_1}, ..., according to its input order. The model must output the reasoning chain in logical order, not in candidate presentation order. Each hop contains the selected \texttt{image\_id}, one or more bounding boxes, and a short natural-language sub-question (or reasoning thought) describing the evidence sought at that hop.

\subsection{Unified Generation with Chain of Evidence}

The final stage synthesizes the selected evidence to produce both an answer and a complete chain of evidence. We model this as a conditional generation problem:

\begin{equation}
P(a, \mathcal{E} | q, \mathcal{D}^{\text{cand}}) =
\prod_{t=1}^{T} P(d_t^*, \mathcal{B}_t, r_t | q, \mathcal{D}^{\text{cand}}, e_{<t}),
\end{equation}
where $r_t$ is the textual sub-query associated with hop $t$. 

\section{Experiment Setup}

We design a comprehensive evaluation protocol to assess CoE's capabilities across two distinct regimes: (1) large-scale multi-hop reasoning on structured web documents (Wiki-CoE), and (2) complex visual understanding on free-form presentation slides (SlideVQA). This dual-dataset approach validates CoE's generalization from standard layouts to scenarios where visual spatial relationships are the primary information carriers.

\subsection{Datasets}

\textbf{Wiki-CoE (Structured Web Layouts).}
As described in Section \ref{sec:wiki_coe}, we utilize our constructed Wiki-CoE benchmark to evaluate pixel-level attribution in a large-scale, open-domain setting. This dataset challenges the model to identify and localize evidence across standard HTML-rendered Wikipedia pages, serving as a testbed for general multi-hop reasoning capabilities.

\textbf{SlideVQA (Complex Visual Layouts).}
To rigorously evaluate CoE's core motivation, handling documents where text extraction is brittle or insufficient, we incorporate the SlideVQA dataset \cite{SlideVQA2023}. SlideVQA consists of 2,619 slide decks (approx. 52k images) with multi-hop questions that require synthesizing information across multiple slides.
Unlike Wikipedia pages, presentation slides feature free-form layouts, diagrams, arrows, and charts where the spatial arrangement is semantically crucial. Traditional OCR engines often fail to preserve the reading order or structural logic of these elements, making this an ideal testbed for our visual-first paradigm.

\subsection{Evaluation Metrics}

We evaluate CoE along three critical dimensions across both datasets:

\textbf{Answer Accuracy.} We employ exact match (\textbf{EM}) to evaluate generated answers, following established multi-hop QA conventions.

\textbf{Evidence Localization Accuracy (Loc-Acc).} In the top-5 candidate setting, localization is counted as correct only when the model selects the correct candidate image for each evidence hop and its predicted bounding box overlaps the ground-truth region. A bounding box match is accepted when IoU $\geq$ 0.3 or the predicted box center falls inside the ground-truth evidence region. Thus Loc-Acc is a joint image-and-box metric rather than a box-only score.

\textbf{Reasoning Chain Accuracy (Chain-Acc).} In the top-5 candidate setting, we verify whether the model selects the correct visual document at each hop and whether the ordered document chain matches the gold reasoning path. We also report joint chain metrics that require both the correct candidate image and a correct evidence box at each hop.

\subsection{Baselines}

We compare CoE against strong baselines representing different paradigms. For a fair comparison under the retriever-agnostic setting, all baselines are provided with the same top-5 candidate documents (parsed as text via OCR for visually heavy datasets like SlideVQA).

\subsubsection{Text-based iRAG} ~\\
\textbf{Strong text-based iRAG baselines:} (1) KiRAG \cite{DBLP:conf/acl/FangMM25} is the recent state of-the-art iRAG method; (2) SEAKR \cite{DBLP:conf/acl/0002QPCHL0L25} is another strong baseline of iRAG.

\textbf{Text-based Attribution Methods:} (1) ALCE-VA-citation \cite{DBLP:conf/emnlp/GaoYYC23} that generates inline citations with document references, adapted to output text-level attributions; (2) IRCOT \cite{DBLP:conf/acl/TrivediBKS23} implementing iterative retrieval with chain-of-thought reasoning but only text-level citations.

\subsubsection{Vision-Language Models} ~\\
GPT-5 and Qwen3-VL-235B evaluated in zero-shot settings with carefully crafted prompts for evidence localization. These baselines assess the inherent capability of proprietary SOTA models without task-specific fine-tuning.

\begin{table*}[t]
\centering
\small
\setlength{\tabcolsep}{4pt}
\begin{tabular}{lcccccc}
\toprule
\multirow{3}{*}{\textbf{Method}} & \multicolumn{3}{c}{\textbf{Wiki-CoE (Web Layouts)}} & \multicolumn{3}{c}{\textbf{SlideVQA (Complex Layouts)}} \\
\cmidrule(lr){2-4} \cmidrule(lr){5-7}
& \textbf{EM} & \textbf{Chain-Acc} & \textbf{Loc-Acc} & \textbf{EM} & \textbf{Chain-Acc} & \textbf{Loc-Acc} \\
\midrule
\multicolumn{7}{l}{\textit{Text-based Attribution Baselines (OCR-based for SlideVQA)}} \\
IRCOT \cite{DBLP:conf/acl/TrivediBKS23} & 57.8 & 54.6 & - & 34.2 & 28.5 & - \\
ALCE-VA-citation \cite{DBLP:conf/emnlp/GaoYYC23} & 58.5 & 56.7 & - & 35.8 & 29.1 & - \\
KiRAG \cite{DBLP:conf/acl/FangMM25} & 60.2 & - & - & 38.1 & - & - \\
SEAKR \cite{DBLP:conf/acl/0002QPCHL0L25} & 60.7 & - & - & 39.4 & - & - \\
\midrule
\multicolumn{7}{l}{\textit{Vision-Language Model Baselines (Zero-shot)}} \\
GPT-5 & 81.2 & 68.1 & 31.7 & 58.5 & 55.4 & 34.1 \\
Qwen3-VL-235B & 78.6 & 66.9 & 7.4 & 58.3 & 51.2 & 6.8 \\
\midrule
\multicolumn{7}{l}{\textit{CoE (Ours)}} \\
CoE-4B (Phase II) & 78.6 & 89.7 & 71.1 & 52.3 & 77.1 & 51.6 \\
CoE-8B (Phase II) & \textbf{82.3} & \textbf{94.4} & \textbf{80.4} & \textbf{58.8} & \textbf{87.5} & \textbf{61.0} \\
\bottomrule
\end{tabular}
\caption{
Main results under the \textbf{top-5 candidate setting} for both Wiki-CoE and SlideVQA datasets.
\textbf{Wiki-CoE} represents structured HTML environments, while \textbf{SlideVQA} represents free-form visual layouts where spatial semantics are critical.
}
\vspace{-3mm}
\label{tab:main_results_combined}
\end{table*}

\subsection{Model Implementation}

We employ Qwen3-VL-8B-Instruct as our primary VLM backbone. For scale analysis, we also report a smaller Qwen3-VL-4B-Instruct variant under the same top-5 candidate evaluation protocol.

Our training follows a two-phase curriculum. Phase I focuses on single-hop evidence localization, establishing visual grounding capabilities with a compact single-image JSON target. Phase II introduces multi-hop evidence-chain generation over top-5 candidate screenshots, warm-starting from the Phase I checkpoint when available. We fine-tune Qwen3-VL with the standard autoregressive language-modeling loss on the assistant JSON response, masking system and user tokens in the loss.
To enhance robustness, we incorporate several training-time augmentation strategies:
    \textbf{1. Spatial augmentation}: We apply geometric perturbations such as random cropping, translation, and aspect-ratio variation to improve robustness to layout shifts, with bounding boxes transformed consistently.
    \textbf{2. Resolution variation}: We expose the model to multiple input resolutions so it can balance global layout understanding with fine-grained OCR readability across documents of different visual density.
    \textbf{3. Evidence permutation}: We perturb the presentation order of evidence or candidate documents while preserving the supervised logical chain order, discouraging positional shortcuts in multi-hop reasoning.

We evaluate CoE in a top-5 candidate setting to decouple evidence reasoning from any specific retriever implementation.

\section{Experimental Results}

\subsection{Main Results}

We present the end-to-end performance comparison on both Wiki-CoE (structured web layouts) and SlideVQA (complex free-form layouts) in Table~\ref{tab:main_results_combined}. The results demonstrate the efficacy of CoE across diverse visual environments.

\textbf{CoE Turns Answers into Verifiable Evidence Chains.}
On the Wiki-CoE benchmark, CoE-8B achieves state-of-the-art performance with 82.3\% EM, 94.4\% Chain-Acc, and 80.4\% Loc-Acc. Strong general VLMs already obtain competitive answer accuracy (e.g., GPT-5 reaches 81.2\% EM and Qwen3-VL-235B reaches 78.6\% EM), but their attribution quality is much weaker. Even CoE-4B, whose EM is comparable to Qwen3-VL-235B, reaches 89.7\% Chain-Acc and 71.1\% Loc-Acc, outperforming GPT-5 by 21.6 and 39.4 points on the two attribution metrics. This contrast shows that answer generation and faithful visual attribution are separable capabilities; high EM alone does not guarantee that a model can identify the ordered evidence documents or point to the supporting regions.

\textbf{Visual-First Approach is Critical for Complex Layouts.}
The same pattern becomes sharper on SlideVQA. General VLMs can still answer many questions correctly: GPT-5 and Qwen3-VL-235B achieve 58.5\% and 58.3\% EM, respectively, nearly matching CoE-8B's 58.8\% EM. However, their evidence-chain and bounding-box grounding remain unreliable. CoE-8B improves over GPT-5 by 32.1 points in Chain-Acc (87.5\% vs. 55.4\%) and 26.9 points in Loc-Acc (61.0\% vs. 34.1\%); compared with Qwen3-VL-235B, the localization gap expands to 54.2 points (61.0\% vs. 6.8\%). CoE-4B shows the same attribution-oriented behavior: despite lower EM than the largest zero-shot VLMs, it still surpasses GPT-5 by 21.7 points in Chain-Acc and 17.5 points in Loc-Acc. This confirms our core hypothesis: visually rich document QA requires not only producing the correct answer, but also grounding that answer in the correct visual reasoning path.

\textbf{General VLMs Lack Precision in Attribution.}
The weakness of zero-shot VLMs is therefore not primarily answer synthesis, but verifiability. Their generated answers can be correct because they recognize salient text, rely on broad parametric knowledge, or infer from partial visual cues, yet their evidence outputs often select the wrong candidate image, omit hops, or fail to place precise bounding boxes. This is particularly problematic for iRAG, where users need to audit \textit{why} an answer is correct. CoE's supervised JSON evidence chains directly optimize the ordered image selection and region-level grounding that generic instruction tuning does not reliably teach.

\textbf{Impact of Model Scale and Training.}
Scaling from 4B to 8B consistently improves both answer accuracy and attribution quality. On Wiki-CoE, CoE-8B improves over CoE-4B by 3.7 points in EM, 4.7 points in Chain-Acc, and 9.3 points in Loc-Acc; on SlideVQA, the gains are 6.5, 10.4, and 9.4 points, respectively. This suggests that task-specific CoE supervision contributes most strongly to evidence-chain construction and precise visual grounding, while larger backbones further improve robustness across both structured web pages and noisy slide content.

\begin{figure*}[t]
    \centering
    \includegraphics[width=0.95\textwidth]{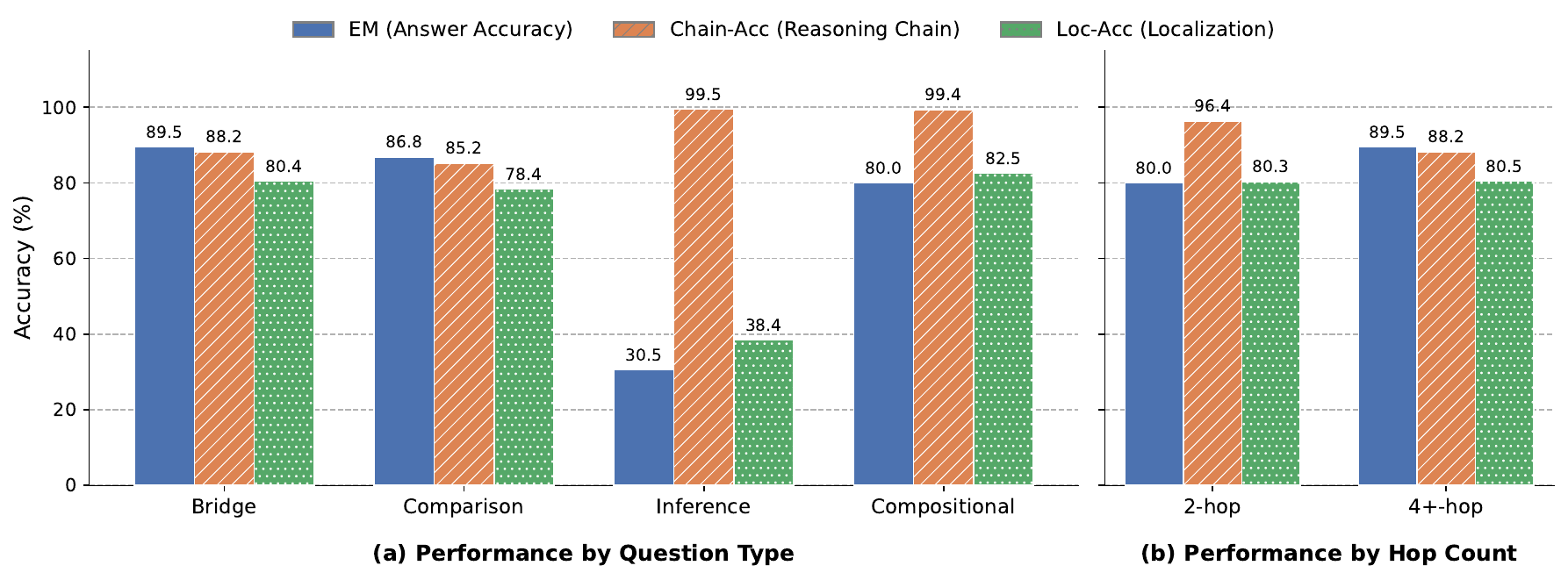}
    \vspace{-3mm}
    \caption{
    CoE-8B performance breakdown by question type and reasoning depth. 
    }
    \label{fig:performance_analysis}
    \vspace{-3mm}
\end{figure*}

\subsection{Performance by Reasoning Type (Wiki-CoE)}
\label{sec:performance_analysis}

To understand how visual modality interacts with logical complexity, we analyze CoE-8B's performance across different reasoning patterns (Figure~\ref{fig:performance_analysis}). Our analysis reveals three distinct behaviors characterizing visual evidence-chain reasoning.

We observe that CoE excels when evidence is explicitly encoded in structured layouts. \textbf{Bridge-comparison} and \textbf{Comparison} questions achieve the strongest answer accuracy (89.5\% and 86.8\% EM, respectively), while \textbf{Compositional} questions obtain the most reliable evidence routing with 99.4\% Chain-Acc and 82.5\% Loc-Acc. This suggests that visual cues, such as table alignments, infobox separators, and parallel list structures, serve as strong inductive biases for the model. For comparison questions specifically, CoE attains 85.2\% Chain-Acc and 78.4\% Loc-Acc, indicating that parallel attributes can be identified reliably, although precisely delineating the supporting regions within dense tables remains more difficult than selecting the correct candidate documents.

A critical divergence appears in \textbf{Inference} questions. CoE almost always identifies the correct evidence documents (99.5\% Chain-Acc), yet EM drops to 30.5\% and Loc-Acc drops to 38.4\%. This substantial gap ($\Delta \approx 61.0\%$ between Chain-Acc and Loc-Acc) highlights a fundamental challenge in visual RAG: \textit{grounding implicit logic}. Unlike explicit fact lookup, inference requires synthesizing unwritten connections. The model can select the relevant source documents but struggles to generate a bounding box around ``reasoning'' that is not explicitly rendered as text, suggesting that current VLMs still treat evidence localization largely as semantic region matching.

\textbf{Reasoning depth} mainly affects ordered document selection rather than final box grounding. Across the full Wiki-CoE test set, 2-hop questions achieve 96.4\% Chain-Acc and 80.3\% Loc-Acc, while 4-hop questions achieve 88.2\% Chain-Acc and 80.5\% Loc-Acc. The 8.2-point chain drop confirms that longer trajectories still introduce error propagation, but the nearly unchanged localization accuracy shows that once the correct evidence documents are selected, CoE can localize supporting regions robustly even in longer chains. Future work should therefore focus on error-correcting mechanisms for long-horizon visual planning.

\subsection{Generalization to Complex Layouts (SlideVQA)}
\label{sec:slidevqa_results}

To further dissect the impact of visual elements, we categorized the SlideVQA test set into three levels of visual complexity based on the density of non-textual elements (charts, diagrams, arrows, and free-form sketches), results are shown in Figure \ref{fig:slide_complexity}:

\begin{figure}[t]
    \centering
    \includegraphics[scale=0.4]{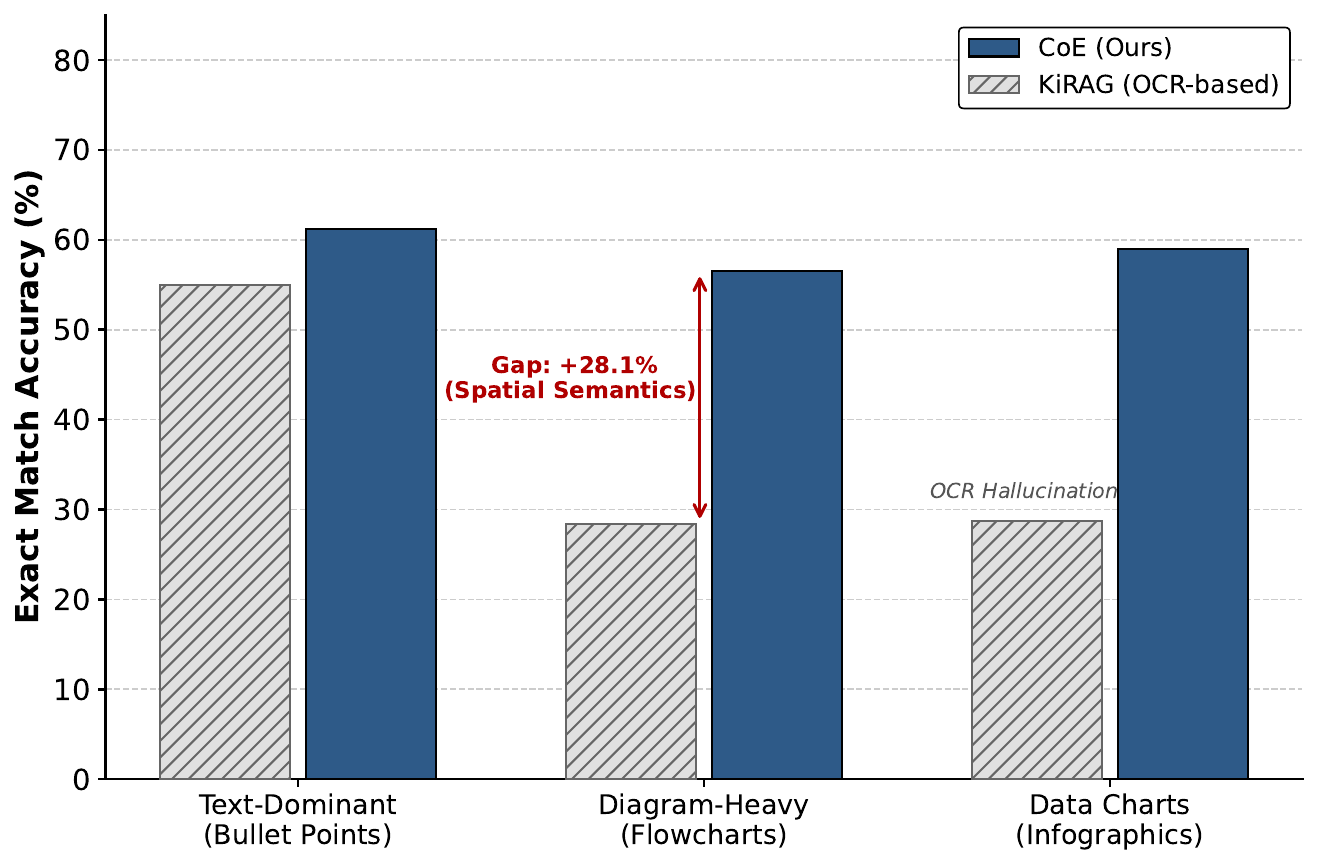}
    \vspace{-9mm}
    \caption{Performance degradation analysis across increasing visual complexity.}
    \label{fig:slide_complexity}
    \vspace{-3mm}
\end{figure}

\textbf{1. Text-Dominant Slides:} On slides consisting primarily of bullet points and headers, the gap between CoE-8B (61.0\% EM) and the strongest OCR baseline (55.0\% EM) is the smallest among all subsets (${\sim}6\%$). The reading order on these slides is generally left-to-right, top-to-bottom, which OCR handles relatively well, yet CoE still benefits from preserving the visual hierarchy of headers and lists.

\textbf{2. Diagram-Heavy Slides:} The divergence becomes extreme on slides featuring flowcharts, organizational charts, and cycle diagrams. OCR-based performance plummets to 28.0\%, as the semantic logic relies entirely on visual connectors (arrows, lines) that are ignored by text extractors. CoE, however, maintains substantially stronger performance at 56.5\%. This \textbf{28.5\% gap} empirically proves that visual attribution is not merely an enhancement but a necessity for interpreting non-linear information structures.

\textbf{3. Data Charts \& Infographics:} For questions requiring data extraction from bar charts or scatter plots, CoE demonstrates precise pixel-level grounding. While text baselines often hallucinate values due to the inability to align axis labels with data bars, CoE's predictions (59.0\% EM) confirm that the model benefits from attending to the specific visual intersection of data points and axes.

\begin{table*}[t]
\centering
\small
\setlength{\tabcolsep}{5pt}
\begin{tabular}{lcccccc}
\toprule
\multirow{2}{*}{\textbf{Configuration}} & \multicolumn{3}{c}{\textbf{Wiki-CoE (Web Layouts)}} & \multicolumn{3}{c}{\textbf{SlideVQA (Complex Layouts)}} \\
\cmidrule(lr){2-4} \cmidrule(lr){5-7}
& \textbf{EM} & \textbf{Chain-Acc} & \textbf{Loc-Acc} & \textbf{EM} & \textbf{Chain-Acc} & \textbf{Loc-Acc} \\
\midrule
\textbf{CoE-8B (Full)} & 82.3 & 94.4 & 80.4 & 58.8 & 87.5 & 61.0 \\
\midrule
\multicolumn{7}{l}{\textit{Training Strategy}} \\
\quad w/o Phase I (Single-hop) & 81.1 & 89.8 & 73.1 & 55.2 & 81.7 & 53.2 \\
\quad w/o Curriculum & 81.7 & 92.1 & 75.9 & 56.6 & 84.2 & 56.0 \\
\midrule
\multicolumn{7}{l}{\textit{Data Augmentation}} \\
\quad w/o Spatial Aug & 82.3 & 92.3 & 75.6 & 54.7 & 82.4 & 54.5 \\
\quad w/o Resolution Var & 80.5 & 92.4 & 76.2 & 56.1 & 83.9 & 57.0 \\
\quad w/o Evidence Perm & 80.2 & 93.5 & 77.6 & 58.1 & 86.1 & 59.1 \\
\quad w/o All Aug & 78.6 & 88.1 & 73.0 & 53.0 & 79.8 & 51.0 \\
\midrule
\multicolumn{7}{l}{\textit{Architecture \& Modality}} \\
\quad Resolution \tiny{(512$\times$512)} & 72.2 & 81.0 & 64.4 & 55.4 & 83.1 & 58.0 \\
\quad Resolution \tiny{(1536$\times$1536)} & \textbf{84.4} & \textbf{95.8} & \textbf{81.9} & \textbf{60.1} & \textbf{88.6} & \textbf{63.0} \\
\quad Text-only Input$^{\ddagger}$ & 56.3 & - & - & 36.5 & - & - \\
\bottomrule
\end{tabular}
\caption{Ablation study results comparing the impact of components across structured (Wiki-CoE) and unstructured (SlideVQA) environments.
$^{\ddagger}$Text-only baseline uses OCR-extracted text, preventing bounding box prediction.}
\label{tab:ablation}
\end{table*}

\subsection{Ablation Studies}
\label{sec:ablation}

To disentangle the contributions of our training strategies, data augmentations, and architectural choices, we conducted a systematic ablation study. The results, presented in Table~\ref{tab:ablation}, offer critical insights into the mechanisms required for effective visual evidence-chain reasoning across both structured (Wiki-CoE) and unstructured (SlideVQA) environments.

\textbf{Curriculum Learning and Task Decomposition.}
Our two-phase training strategy proves essential for mastering the complex dependency between evidence localization and reasoning. 
Removing Phase I (Single-hop) hurts both datasets, with EM dropping by 1.2 points on Wiki-CoE and 3.6 points on SlideVQA. The effect is even clearer on attribution: Wiki-CoE Loc-Acc drops from 80.4\% to 73.1\%, and SlideVQA Loc-Acc drops from 61.0\% to 53.2\%. This confirms that the ability to ground visual evidence in isolation is a strict prerequisite for multi-hop reasoning; without this foundational capability, the model struggles to build accurate evidence chains.
Crucially, the \textit{w/o Curriculum} setting, where single-hop and multi-hop data are mixed during training, also underperforms the full model on attribution (e.g., 75.9\% vs. 80.4\% Loc-Acc on Wiki-CoE and 56.0\% vs. 61.0\% on SlideVQA). While mixing data is superior to omitting single-hop training entirely, it fails to achieve optimal grounding. This suggests that simultaneously optimizing for basic localization and complex reasoning introduces optimization interference. The curriculum strategy effectively disentangles these tasks, allowing the model to stabilize its visual grounding capabilities before tackling higher-order multi-hop candidate reasoning.

\textbf{Visual Robustness via Augmentation.}
The impact of data augmentation reveals a divergence between structured and free-form layouts. On Wiki-CoE, removing spatial augmentation leaves EM nearly unchanged but reduces Chain-Acc and Loc-Acc by 2.1 and 4.8 points, respectively; on SlideVQA, the same ablation causes broader degradation, including a 4.1-point EM drop and a 6.5-point Loc-Acc drop. This disparity underscores the nature of the data: Wikipedia pages follow standardized HTML/CSS templates, allowing the model to rely partially on position biases. In contrast, presentation slides exhibit extreme spatial variability. Spatial augmentation forces the model to learn geometric invariance, ensuring that reasoning relies on relative visual semantics rather than absolute coordinates. Resolution variation provides a complementary form of robustness: removing it lowers Wiki-CoE by 1.8 EM, 2.0 Chain-Acc, and 4.2 Loc-Acc points, and lowers SlideVQA by 2.7 EM, 3.6 Chain-Acc, and 4.0 Loc-Acc points. The consistent attribution drop indicates that multi-scale exposure is not merely an OCR convenience; it teaches the model to preserve evidence grounding across documents whose relevant cues range from dense Wikipedia text to large slide graphics. Evidence permutation further discourages positional shortcuts by decoupling the presentation order from the supervised logical reasoning order.

\textbf{Resolution and Modality Necessity.}
Our architectural ablations validate the fundamental premise of CoE. Reducing input resolution to $512 \times 512$ causes a severe collapse on Wiki-CoE, with EM, Chain-Acc, and Loc-Acc dropping by 10.1, 13.4, and 16.0 points, respectively. This sensitivity is expected for rendered web pages, where evidence often appears as small-font text inside infoboxes, tables, and densely packed paragraphs. SlideVQA is less damaged by this low-resolution setting (3.4 EM, 4.4 Chain-Acc, and 3.0 Loc-Acc point drops), likely because many slides contain larger visual objects and shorter text spans, but the degradation remains consistent across all metrics. Increasing the resolution to $1536 \times 1536$ improves both benchmarks beyond the default setting: Wiki-CoE gains 2.1 EM, 1.4 Chain-Acc, and 1.5 Loc-Acc points, while SlideVQA gains 1.3 EM, 1.1 Chain-Acc, and 2.0 Loc-Acc points. These results show that high-resolution visual input benefits both answer generation and verifiable grounding, especially when evidence is encoded in fine-grained typography or precise visual alignment. Most critically, the \textit{Text-only Input} baseline serves as a lower bound, exhibiting a catastrophic performance gap (26.0 points on Wiki-CoE and 22.3 points on SlideVQA). This empirically proves that for complex documents, visual structure is not merely supplementary context but the primary carrier of logical information, which is inevitably lost in text-only processing. These findings collectively validate that our visual-first architecture, coupled with the curriculum training and spatial and resolution augmentations, is indispensable for robust multi-hop attribution.

\begin{table}[t]
\centering
\small
\setlength{\tabcolsep}{1.3mm}{
\begin{tabular}{lccccc}
\toprule
\textbf{Method} & \textbf{Params} & \textbf{Latency} & \textbf{Memory} & \textbf{FLOPs} \\
\midrule
IRCOT (text) & 7B & 3.2s & 14GB & 1.0$\times$ \\
ALCE-VA (text) & 7B & 2.8s & 14GB & 1.0$\times$ \\
\midrule
GPT-5 & - & 13.7s$^{*}$ & - & - \\
Qwen3-VL & - & 10.4s$^{*}$ & - & - \\
\midrule
CoE-4B & 4B & 4.1s & 18GB & 1.4$\times$ \\
CoE-8B & 8B & 5.6s & 28GB & 2.1$\times$ \\
CoE-8B-4bit & 8B & 4.3s & 16GB & 1.8$\times$ \\
\bottomrule
\end{tabular}
}
\caption{Computational efficiency comparison (per question, averaged over 3 hops). Latency measured on A800 GPU. Memory includes model weights and activation. FLOPs normalized to IRCOT. $^{*}$API latency, includes network overhead.}
\label{tab:efficiency}
\end{table}

\begin{figure*}[t]
    \centering
    \includegraphics[scale=0.35]{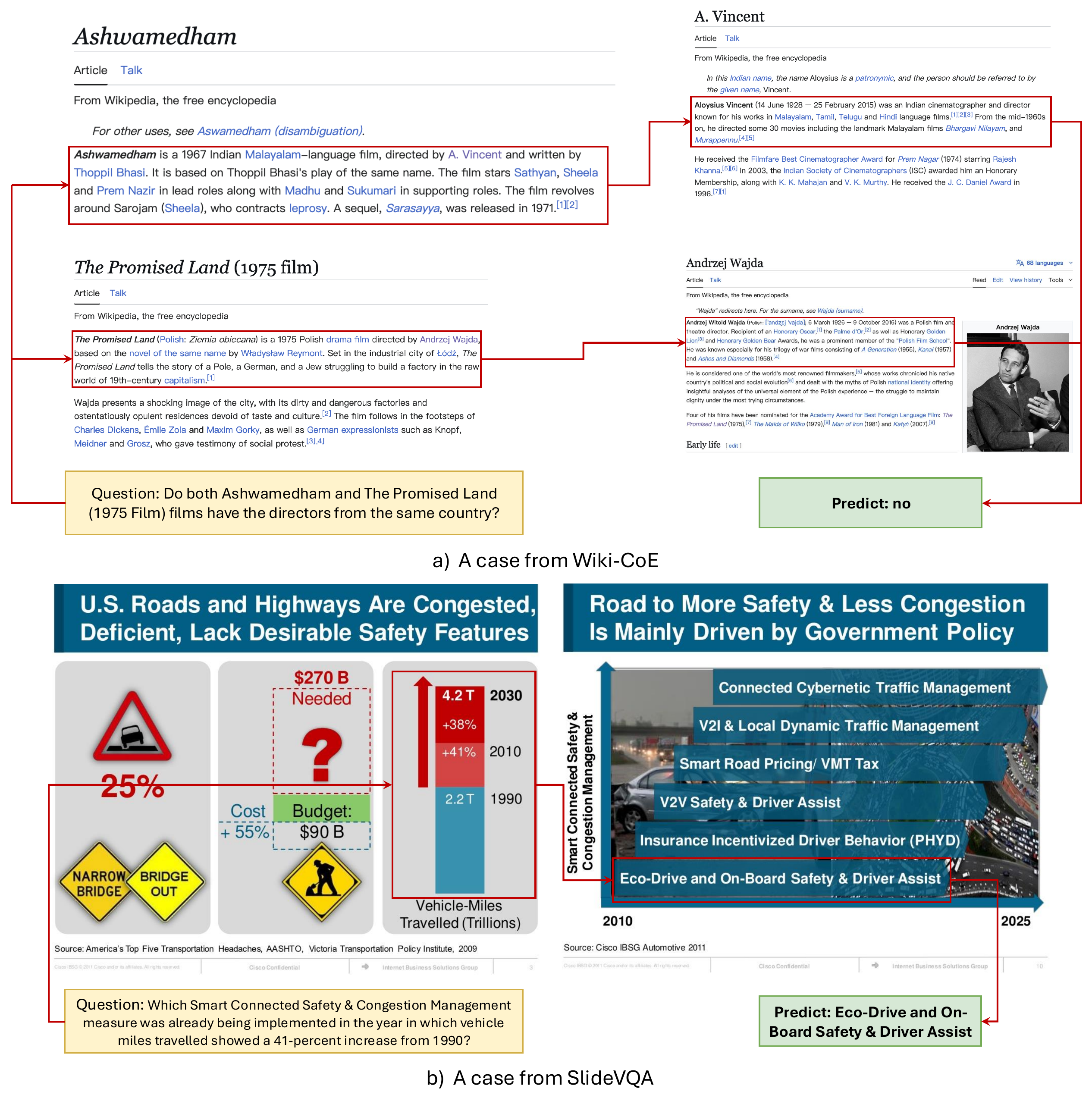}
    \vspace{-3mm}
    \caption{Case studies demonstrating CoE's visual attribution. 
    }
    \label{fig:case_studies}
\end{figure*}

\subsection{Computational Efficiency Analysis}

Table~\ref{tab:efficiency} compares computational costs across methods, addressing practical deployment considerations.

Despite processing visual inputs, CoE-8B adds acceptable computational overhead compared to text-based methods (5.6s vs 3.2s), while providing substantially richer attribution. This modest increase stems from efficient vision encoder design and the fact that candidate screenshots are processed together in the VLM context, with subsequent reasoning operating on compressed visual embeddings.

4-bit quantization reduces CoE-8B's memory footprint from 28GB to 16GB (43\% reduction) with minimal accuracy loss ($<1\%$ EM drop, not shown in tables). This brings CoE within reach of consumer GPUs, enabling broader deployment. Latency decreases to 4.3s, approaching text-based methods while maintaining visual attribution capabilities.

\subsection{Qualitative Analysis}
\label{sec:case_studies}

To intuitively demonstrate CoE's reasoning mechanisms and the necessity of visual grounding, we visualize two representative inference trajectories in Figure~\ref{fig:case_studies}.

Figure~\ref{fig:case_studies}(a) illustrates a complex comparison query requiring evidence from four distinct web pages. The model successfully decomposes the query, first selecting film entries to identify directors, and subsequently selecting their biographies to determine nationalities. Unlike text-based iRAG which generates a final answer that is often hard to verify, CoE provides an explicit \textit{visual audit trail}. By placing bounding boxes on the specific infobox rows (e.g., identifying ``A. Vincent'' and ``Andrzej Wajda''), CoE proves that it has correctly disambiguated the entities rather than hallucinating connections. This pixel-level grounding allows users to instantly validate intermediate reasoning steps, addressing the ``verification bottleneck'' in long-chain inference.

Figure~\ref{fig:case_studies}(b) highlights the critical advantage of CoE in scenarios where text linearization fails. The query requires correlating a statistical trend (``41\% increase'') with a specific timeline of safety measures.
Standard OCR baselines fail here because the relationship between the bar height, the percentage label, and the year (``2010'') is purely spatial, not lexical. CoE correctly aligns these visual elements to identify the target year. Furthermore, it interprets the graphical arrows in the timeline to determine which measures were ``already implemented'', a logical dependency encoded in the layout flow rather than text. This confirms that CoE does not merely read text from images, but actively reasons over the \textit{visual syntax} of the document, capturing semantic cues that are inevitably lost in text-only processing.
\balance
\section{Conclusion}
In this work, we introduce \textbf{Chain of Evidence (CoE)}, a paradigm shift in Iterative Retrieval-Augmented Generation that transitions from brittle text parsing to robust visual reasoning. By grounding multi-hop inference directly in document screenshots, CoE addresses two fundamental limitations of existing systems: the loss of semantic layout information and the opacity of source attribution. Our comprehensive evaluation on the newly constructed \textbf{Wiki-CoE} benchmark and the complex \textbf{SlideVQA} dataset reveals a critical insight: visual modality is not merely a supplement for interpretability but a necessity for reasoning over visually rich knowledge where spatial logic supersedes linear text.
Empirically, our fine-tuned VLM demonstrates that pixel-level grounding significantly outperforms text-based baselines, providing a visual audit trail that effectively resolves the verification bottleneck. By proving that visual grounding serves as an effective inductive bias for complex reasoning, our work challenges the prevailing text-centric view of information retrieval. As RAG systems are increasingly deployed in high-stakes domains, CoE offers a blueprint for the next generation of verifiable AI. Future work will explore extending this unified visual framework to handle heterogeneous web content, such as dynamic video frames and interactive app interfaces, paving the way for truly universal autonomous agents.
\bibliographystyle{ACM-Reference-Format}
\bibliography{sample-base}

\end{document}